\documentclass{article}
\usepackage[numbers,compress]{natbib}
\usepackage{amssymb}
\usepackage{multirow}
\usepackage{amsmath}
\usepackage{pifont}

\usepackage{array}

\usepackage[preprint]{corl_2025} 
\usepackage{amsmath} 
\usepackage{amssymb}  
\usepackage{balance}
\usepackage{graphicx}
\usepackage{times}
\usepackage{dsfont}
\usepackage{multicol}
\usepackage{siunitx}
\usepackage{xspace}
\usepackage{xurl}
\usepackage{capt-of}

\usepackage[usenames,dvipsnames,table,xcdraw]{xcolor}
\usepackage{wrapfig,lipsum,booktabs}
\usepackage{booktabs}
\definecolor{lightgray}{gray}{0.9}
\usepackage{wrapfig}
\usepackage{tikz}
\usepackage{bbding}
\usepackage{cancel}
\usepackage{cleveref}
\usepackage[usenames,dvipsnames]{xcolor}
\usepackage{enumitem}
\usepackage{subcaption}
\definecolor{mydarkblue}{rgb}{0,0.08,0.45}
\definecolor{mydarkgreen}{RGB}{0, 139, 69}
\definecolor{mygreen2}{RGB}{0 205 0}
\definecolor{mybrown}{RGB}{139 69 19}
\definecolor{myred}{RGB}{188, 0, 0}
\definecolor{myblue}{RGB}{0, 0, 198}
\definecolor{ourcolor}{RGB}{118,10,0}
\definecolor{ourglowcolor}{RGB}{230,190,0}
\definecolor{baselinecolor}{RGB}{225,225,0} 
\definecolor{baselineglowcolor}{RGB}{200,0,255}

\definecolor{boxblue}{RGB}{79,173,234}
\definecolor{boxgreen}{RGB}{159,206,99}
\hypersetup{
	colorlinks=true,
	linkcolor=blue,
	urlcolor=magenta,
	citecolor=mygreen2,
}
\title{Hold My Beer: Learning Gentle Humanoid Locomotion and End-Effector Stabilization Control}

\newcommand{\method}{\textcolor{myred}{SoFTA}\xspace}
\usepackage{graphicx}
\begin{document}

\maketitle
\vspace{-50pt}
\begin{center}
\textbf{Yitang Li\footnotemark[1], Yuanhang Zhang, Wenli Xiao, Chaoyi Pan,} \\
\textbf{Haoyang Weng\footnotemark[1], Guanqi He, Tairan He, Guanya Shi} \\
Carnegie Mellon University \\
Website: \href{https://lecar-lab.github.io/SoFTA/}{\texttt{https://lecar-lab.github.io/SoFTA/}} \\
Code: \href{https://github.com/LeCAR-Lab/SoFTA}{\texttt{https://github.com/LeCAR-Lab/SoFTA}}
\end{center}
\footnotetext[1]{This work was done when the author was a visiting intern at CMU.}
\vspace{5pt}


\begin{figure}[h]
    \centering
    \vspace{-15pt}
    \includegraphics[width=0.81\linewidth]{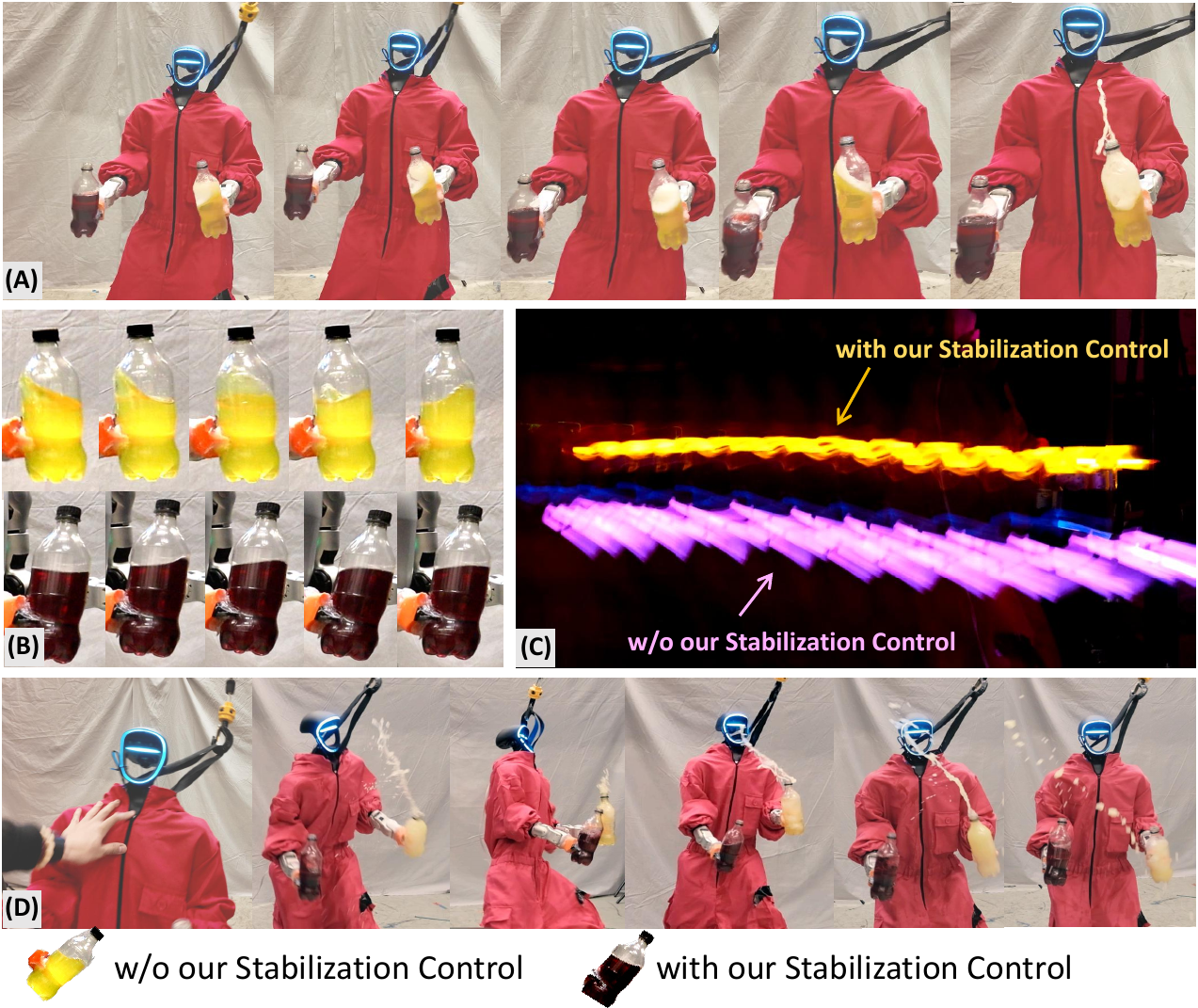}
    \vspace{-4pt}
    \caption{Learning Gentle Humanoid Locomotion and End-Effector Stabilization Control with \method{}:
    \textbf{(A)} Carrying bottles of drink during a $1$m/s large-step walk. 
    \textbf{(B)} Liquid surface when the robot is tapping in place. 
    \textbf{(C)} Long-exposure photo showing the robot holding a glow stick walks forward.
    \textbf{(D)} \method{} keeps the drink from spilling, even after a fierce push. 
}
    \label{fig:teaser}
    \vspace{-12pt}
\end{figure}

\begin{abstract} 
\emph{Can your humanoid walk up and hand you a full cup of beer—without spilling a drop?}
While humanoids are increasingly featured in flashy demos—dancing, delivering packages, traversing rough terrain—fine-grained control during locomotion remains a significant challenge. In particular, stabilizing a filled end-effector (EE) while walking is far from solved, due to a fundamental mismatch in task characteristics: locomotion demands slow-timescale, robust control, whereas EE stabilization requires rapid, high-precision corrections.
To address this, we propose \method, a Slow-Fast Two-Agent framework that decouples upper-body and lower-body control into separate agents operating at different frequencies and with distinct rewards. 
This temporal and objective separation mitigates policy interference and enables coordinated whole-body behavior.
\method executes upper-body actions at 100 Hz for precise EE control and lower-body actions at 50 Hz for robust gait. It reduces EE acceleration by 2-5$\times$ to baselines and performs much closer to human-level stability, enabling delicate tasks such as carrying nearly full cups, capturing steady video during locomotion, and disturbance rejection with EE stability. 
\end{abstract}





\section{Introduction}
\vspace{-4pt}

Humanoid robots are designed to operate in human-centric environments, with general-purpose structures that make them well-suited for diverse tasks. 
Recent advances in locomotion~\cite{liao2024berkeley, li2024reinforcement, radosavovic2024real, radosavovic2402humanoid, gu2024advancing, zhang2024whole, long2024learning, zhuang2024humanoid, wang2025beamdojolearningagilehumanoid, ren2025vbcomlearningvisionblindcomposite, xie2025humanoidwholebodylocomotionnarrow} and manipulation~\cite{qiu2025humanoidpolicyhuman, lin2025simtorealreinforcementlearningvisionbased, li2024okamiteachinghumanoidrobots, atar2025humanoidshospitalstechnicalstudy, mani} have pushed humanoid performance toward human-level capabilities~\cite{gu2025humanoidlocomotionmanipulationcurrent}.
However, one critical capability remains underexplored: fine-grained end-effector (EE) stabilization during locomotion.
This capability is essential for safe and precise physical interaction with objects—such as handing over a cup of water or recording stable video—yet current humanoids fall short. For instance, the default Unitree G1 controller yields average EE accelerations around 5m/s$^2$ when tapping in place —over 10× higher than human levels—leading to excessive shaking and making delicate tasks infeasible.

We identified such a fundamental performance gap stemming from the disparities in task characteristics between EE stabilization and locomotion, both in terms of task objectives and dynamics. At the \emph{task objective} level, locomotion requires traversability, which naturally introduces non-quasi-static dynamics. In contrast, EE stabilization requires a base with minimal motion to maintain precision. At the \emph{dynamics} level, lower body locomotion operates with ``slow" dynamics meaning it can only be controlled through discrete foot contacts in relatively large time scales. The nature of ground contacts makes it more susceptible to the sim-to-real gap, demanding greater robustness against noise and disturbances. On the other hand, EE control involves ``fast'' dynamics, with fully actuated and more controllable arms to produce continuous torques, allowing for fast and precise corrections.

To bridge the gap, we propose \textbf{\method}—a \textbf{\textcolor{myred}{S}}l\textbf{\textcolor{myred}{o}}w-\textbf{\textcolor{myred}{F}}ast \textbf{\textcolor{myred}{T}}wo-\textbf{\textcolor{myred}{A}}gent reinforcement learning (RL) framework that decouples the action and value spaces of the upper and lower body. This design enables different execution frequencies and reward structures: the upper-body agent acts at high frequency for precise EE control with compensation behavior, while the lower-body agent prioritizes robust locomotion at a slower frequency. \method{} facilitates stable training and whole-body coordination by this decoupling, resulting in fast and accurate EE control alongside robust locomotion. Like shown in Figure~\ref{fig:teaser}, our system achieves a 50--80\% reduction in EE acceleration over baselines. \method can achieve EE acceleration less than $2$m/s\textsuperscript{2} in diverse locomotion, which is much closer to human-level stability, enabling tasks like serving coffee or stable video recording. Our key contributions are:

\begin{itemize}[leftmargin=*]
\item We introduce \method, a novel slow-fast two-agent RL framework that decouples control for locomotion and EE stabilization in both temporal and task objective space, enabling robust locomotion and precise, stable EE control through frequency separation and task-specific reward design.
\item We demonstrate real-world deployment of \method on a Unitree G1 and Booster T1 humanoids, enabling tasks such as walking while carrying liquids or recording stable first-person videos. 
\item Extensive experiments are conduct in both simulation and real-world with in-depth analysis across control frequencies, showing that \method can effectively stabilizes the end-effector during locomotion through its frequency design.
\end{itemize}

\section{Related Work}
\vspace{-4pt}
\label{sec:citations}

\paragraph{Learning-based Humanoid Control}

Recent advances in learning-based whole-body control have enabled humanoid robots to acquire a wide range of skills in simulation. Efforts such as domain randomization and system identification to better align simulation with real-world dynamics~\cite{peng2018sim, tan2018sim, chen2022understandingdomainrandomizationsimtoreal, asap, sobanbabu2025sampling} have proven to be effective for sim-to-real transfer of humanoid policies. These capabilities span robust locomotion~\cite{li2019using, xie2020learning, li2021reinforcement, liao2024berkeley, li2024reinforcement, radosavovic2024real, radosavovic2402humanoid, gu2024advancing, zhang2024whole, long2024learning, zhuang2024humanoid, wang2025beamdojolearningagilehumanoid, ren2025vbcomlearningvisionblindcomposite, xie2025humanoidwholebodylocomotionnarrow,he2024hover}, advanced manipulation~\cite{qiu2025humanoidpolicyhuman, lin2025simtorealreinforcementlearningvisionbased, li2024okamiteachinghumanoidrobots, atar2025humanoidshospitalstechnicalstudy, mani} and integrated loco-manipulation behaviors~\cite{he2024learning, lu2024mobile, fu2024humanplus, he2024omnih2o, ben2025homiehumanoidlocomanipulationisomorphic, shi2025toddlerbotopensourcemlcompatiblehumanoid, shi2025adversariallocomotionmotionimitation, gu2025humanoidlocomotionmanipulationcurrent}.
Despite these promising developments, relatively little attention has been paid to achieving precise and stable EE control, which is essential for fine-grained humanoid loco-manipulation. In this work, we focus on enabling humanoid robots maintain end-effector stability during locomotion.

\paragraph{End-Effector Control for Mobile Manipulators} Stabilizing EE during motion is crucial for mobile manipulation. Prior work predominantly focuses on wheeled or aerial robots, where model-based approaches unify base and arm control through optimization~\cite{hyq_with_arm, sentis2005synthesis, ferrolho2023roloma, optimizingdynamictraj, wheelee, mobilemani1, osman2021endeffectorstabilization10dofmobile, Minniti_2019,guo2024flying,he2025flying}, but they rely on accurate dynamics models and predefined contact schedules, limiting their scalability to complex humanoid systems. Hybrid approaches~\cite{ma2022combining,liu2024visual,roboduet} combine learned locomotion with planned arms but often freeze the base, reducing coordination. Joint learning~\cite{fu2022deepwholebodycontrollearning,portela2025wholebodyendeffectorposetracking} improves tracking, mostly on non-humanoid robots. In contrast, we present the first method to achieve fine-grained end-effector stabilization during dynamic humanoid locomotion.

\paragraph{Humanoid Policy Architecture}
To enable effective humanoid policy learning, researchers have explored various architectural designs~\cite{shi2025adversariallocomotionmotionimitation, lu2025mobiletelevisionpredictivemotionpriors,zhang2025falconlearningforceadaptivehumanoid, guo2023decentralizedmotorskilllearning, Pang_2023}. Single-agent whole-body policies~\cite{asap, zhang2024wococolearningwholebodyhumanoid} offer flexibility for complex tasks but are challenged by high-dimensional state-action spaces. Inspired by MARL~\cite{zhang2021multiagentreinforcementlearningselective}, recent work decouples policies to simplify training. Multi-critic methods~\cite{Zhuang2025EmbraceCH, huang2025learninghumanoidstandingupcontrol} handle reward conflicts, while decentralized control~\cite{guo2023decentralizedmotorskilllearning} assigns body parts to separate controllers. Others~\cite{shi2025adversariallocomotionmotionimitation, lu2025mobiletelevisionpredictivemotionpriors, zhang2025falconlearningforceadaptivehumanoid} split locomotion and manipulation, enabling diverse behaviors. Still, few leverage the underlying dynamics of humanoid robots for effective architecture designs. Our method extends this idea to both frequency and upper-lower body decoupling by capturing the fundamental disparity in task characteristics, achieving stable EE control and robust locomotion.

\section{\method for Learning Stable End-effector Control and  Robust Locomotion}
\vspace{-4pt}

\subsection{Problem Statement}
\vspace{-4pt}
\paragraph{Observations and Actions.} We aim to control a humanoid robot to stabilize its end-effector at target positions while also following locomotion commands. We formulate the problem as a goal-conditioned RL task, where the policy \(\pi(s_t^{\text{prop}}, s_t^{\text{goal}})\) is trained to output an action \(a_t \in \mathbb{R}^{27}\), representing target joint positions. The proprioceptive input \(s_t^{\text{prop}}\) includes a 5-step history of joint positions \(q_t \in \mathbb{R}^{27}\), joint velocities \(\dot{q}_t \in \mathbb{R}^{27}\), root angular velocities \(\omega_t^{\text{root}} \in \mathbb{R}^3\), projected gravity vectors \(g_t \in \mathbb{R}^3\), and past actions \(a_t \in \mathbb{R}^{27}\). The goal state \(s_t^{\text{goal}}\) contains target root linear velocity \(v_t^{\text{goal}} \in \mathbb{R}^2\), target yaw angular velocity \(\omega_t^{\text{goal}} \in \mathbb{R}\),  desired base heading \(h_t^{\text{goal}} \in \mathbb{R}\), \(c_t^{\text{goal}} \in \mathbb{R}^2\) (with a binary stand/walk command and a gait frequency), and \(c_t^{\text{EE}} \in \mathbb{R}^{5 \times n}\) encodes the EE command. Here, \(n\) denotes the number of potential end-effectors, each with a 5-dimensional command specifying whether it is activated for stabilization, the \(x\)- and \(y\)-coordinates in the local frame, the \(z\)-coordinate in the global frame, and the tracking tolerance \(\sigma\). 
\vspace{-3pt}
\paragraph{Reward Formulation for Stable EE control} We use PPO~\cite{schulman2017proximal} to maximize the cumulative discounted reward \(\mathbb{E}\left[\sum_{t=1}^T \gamma^{t-1} r_t\right]\). Several rewards \(r_t\) are defined to achieve stable end-effector control: 1) penalizing high linear/angular acceleration, \( r_{\text{acc}} = -\| \ddot{p}_{\text{EE}} \|_2^2 \), \( r_{\text{ang-acc}} = -\| \dot{\omega}_{\text{EE}} \|_2^2 \); 2) encouraging near-zero linear/angular acceleration, \( r_{\text{zero-acc}} = \exp\left( -\lambda_{\text{acc}} \| \ddot{p}_{\text{EE}} \|_2^2 \right) \), \( r_{\text{zero-ang-acc}} = \exp\left( -\lambda_{\text{ang-acc}} \| \dot{\omega}_{\text{EE}} \|_2^2 \right) \); 3) penalizing gravity tilt in the end-effector frame, \( r_{\text{grav-xy}} = -\bigl\|\mathbf{P}_{xy}(R_{\text{EE}}^T \mathbf{g})\bigr\|_2^2 \). \(\ddot{p}_{\mathrm{EE}} \in \mathbb{R}^3\) is the linear acceleration, \(\dot{\omega}_{\mathrm{EE}} \in \mathbb{R}^3\) is the angular acceleration, \(\lambda_{\mathrm{acc}}, \lambda_{\mathrm{ang\!-\!acc}} > 0\) are exponential reward scale factors, \(R_{\mathrm{EE}} \in \mathrm{SO}(3)\) is the rotation matrix, \(\mathbf{g}\) is the gravity vector, and \(\mathbf{P}_{xy}(\cdot): \mathbb{R}^3 \to \mathbb{R}^2\) projects onto the \(xy\)-plane.

\vspace{-3pt}
\paragraph{Task Characteristics.}
Stable End-Effector Control and Robust Locomotion are fundamentally different tasks in both their task objectives and dynamics. 

At the \emph{task objective level}, end-effector control demands extreme stability, requiring the base to remain as static as possible, while locomotion must accommodate varying gaits and momentum changes. Precise end-effector control benefit from sharp, fine-grained, continuous rewards, whereas locomotion favors long-horizon, robustness-focused rewards. Given these differences, using a single critic to aggregate all reward signals may not be the most effective way.

At the \emph{dynamics level}, locomotion is governed by discrete ground contact forces and exhibits ``slower" dynamics due to its long time-scales. In contrast, the upper body has a ``faster" dynamics, and is often more controllable by fully actuated arms, affording more aggressive and faster control strategies. Given that higher control frequencies tend to increase sensitivity and exacerbate the sim-to-real gap~\cite{rajeswaran2017epoptlearningrobustneural, Hwangbo_2019, yang2019dataefficientreinforcementlearning, tan2018sim}, while lower frequencies are less precise but more deployable and robust, it is advantageous to modulate control rates accordingly.  

\begin{figure}
    \centering
    \includegraphics[width=0.85\linewidth]{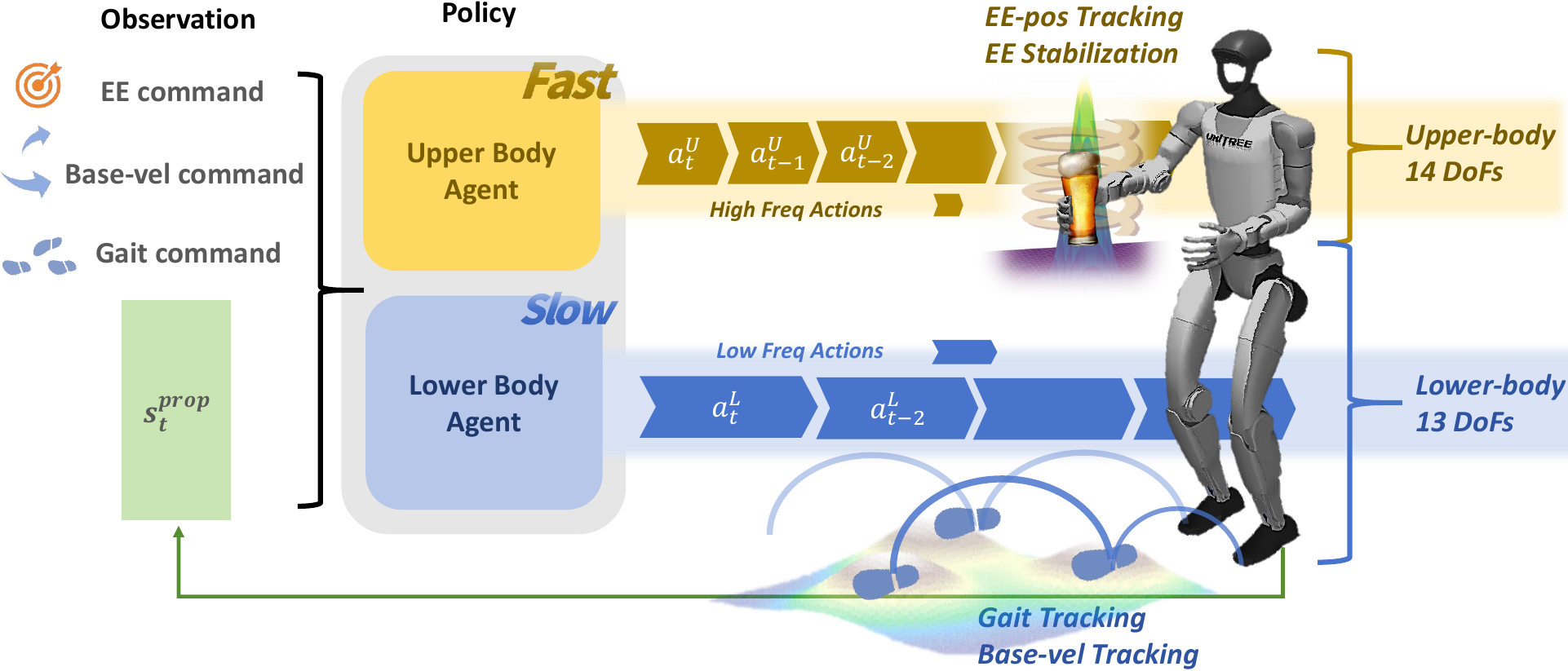}
\caption{Overview of the \method framework: The framework employs two distinct agents that share the same observation but act within separate action spaces at different rates, targeting two fundamentally different task: stable end-effector control and robust locomotion. Stable end-effector control requires a sharp reward landscape and rapid upper-body actions for precise manipulation, whereas robust locomotion focuses on maintaining robustness under gait rewards.}

    \label{fig2}
\end{figure}

\subsection{\method: Slow-Fast Two-Agent Framework.}
\vspace{-3pt}

\paragraph{Slow-Fast Two-Agent Framework Design.} Given these distinct task characteristics, we propose \method, a two-agent framework in which each agent independently controls a disjoint subset of the robot's degrees of freedom at different control frequencies (Figure~\ref{fig2}). Both agents in \method share the full-body observation to facilitate coordinated behavior while allowing each agent to specialize. Specifically, the upper-body agent operates at a high frequency to control 14 arm joints, enabling precise and rapid adjustments for end-effector stability while the lower-body agent runs at a lower frequency, managing the legs and waist to ensure stable locomotion and balance. This asymmetric control frequency matches the longer characteristic timescale of gait cycles compared to the fast, precision motion required for stabilization tasks.

\vspace{-3pt}
\paragraph{Training \method with Separate Reward Groups.} 
 
Due to the differing control dynamics and timescales of upper-body and lower-body tasks, their reward signals are inherently heterogeneous, which can lead to interference and suboptimal learning.
To improve credit assignments~\cite{sunehag2017valuedecompositionnetworkscooperativemultiagent, ma1, ma2, Yarahmadi2023BankruptcyevolutionaryGB}, we decompose the overall reward into two semantically aligned components, each tailored to the respective PPO agent.
This decomposition provides more targeted feedback, preventing overloading of any agent and promoting fair cooperation. To further encourage collaborative behavior and sustained task execution, we include the termination reward in both reward streams. While both agents share the same observation space, they operate with separate actor and critic networks and do not share parameters.
More details are summarized in Appendix~\ref{appendix:training}.

\section{Experimental Results}
\label{sec:result}
\begin{table*}[t]
\small
\vspace{-2mm}
\centering
\resizebox{\linewidth}{!}{%
\begingroup
\setlength{\tabcolsep}{2pt}
\renewcommand{\arraystretch}{0.8}
\begin{tabular}{llcccccccc}
\toprule
\multicolumn{2}{c}{Simulation Results (Isaac Gym)} 
  & \multicolumn{2}{c}{Acc (m/s\textsuperscript{2}) \textcolor{myred}{\pmb{$\downarrow$}}}
  & \multicolumn{2}{c}{AngAcc (rad/s\textsuperscript{2}) \textcolor{myred}{\pmb{$\downarrow$}}} 
  & \multicolumn{2}{c}{Acc-Z (m/s\textsuperscript{2}) \textcolor{myred}{\pmb{$\downarrow$}}} 
  & \multicolumn{2}{c}{Grav-XY (m/s\textsuperscript{2}) \textcolor{myred}{\pmb{$\downarrow$}}} \\
\cmidrule(r){3-4} \cmidrule(r){5-6} \cmidrule(r){7-8} \cmidrule(r){9-10}
Task & Method 
  &  \phantom{0} mean  & \phantom{0} max \phantom{0} 
  & mean & max 
  &  \phantom{0}mean  &  \phantom{0} max  \phantom{0}
  & mean & max \\
\midrule
\multirow{3}{*}{Tapping} 
  & Lower-body RL + IK       & 1.83{\tiny\(\pm\textcolor{gray}{0.05}\)} & 4.92{\tiny\(\pm\textcolor{gray}{0.18}\)} & 10.3{\tiny\(\pm\textcolor{gray}{0.4}\)} & 37.8{\tiny\(\pm\textcolor{gray}{0.8}\)} & 0.77{\tiny\(\pm\textcolor{gray}{0.02}\)} & 3.46{\tiny\(\pm\textcolor{gray}{0.08}\)} & 0.15{\tiny\(\pm\textcolor{gray}{0.01}\)} & 0.53{\tiny\(\pm\textcolor{gray}{0.02}\)} \\
  & Whole-body RL            & 1.29{\tiny\(\pm\textcolor{gray}{0.03}\)} & 4.24{\tiny\(\pm\textcolor{gray}{0.14}\)} & 9.76{\tiny\(\pm\textcolor{gray}{0.25}\)} & \textcolor{myred}{\textbf{31.7}}{\tiny\(\pm\textcolor{gray}{0.7}\)} & 0.58{\tiny\(\pm\textcolor{gray}{0.01}\)} & 1.71{\tiny\(\pm\textcolor{gray}{0.05}\)} & \textcolor{myred}{\textbf{0.09}}{\tiny\(\pm\textcolor{gray}{0.01}\)} & 0.46{\tiny\(\pm\textcolor{gray}{0.02}\)} \\
  & \method                     & \textcolor{myred}{\textbf{1.08}}{\tiny\(\pm\textcolor{gray}{0.03}\)} & \textcolor{myred}{\textbf{3.45}}{\tiny\(\pm\textcolor{gray}{0.11}\)} & \textcolor{myred}{\textbf{8.10}}{\tiny\(\pm\textcolor{gray}{0.20}\)} & 32.1{\tiny\(\pm\textcolor{gray}{0.7}\)} & \textcolor{myred}{\textbf{0.44}}{\tiny\(\pm\textcolor{gray}{0.01}\)} & \textcolor{myred}{\textbf{0.87}}{\tiny\(\pm\textcolor{gray}{0.04}\)} & 0.11{\tiny\(\pm\textcolor{gray}{0.01}\)} & \textcolor{myred}{\textbf{0.44}}{\tiny\(\pm\textcolor{gray}{0.01}\)} \\

\midrule
\multirow{3}{*}{RandCommand} 
  & Lower-body RL + IK       & 3.17{\tiny\(\pm\textcolor{gray}{0.07}\)} & 6.51{\tiny\(\pm\textcolor{gray}{0.21}\)} & 15.5{\tiny\(\pm\textcolor{gray}{0.6}\)} & 53.4{\tiny\(\pm\textcolor{gray}{1.2}\)} & 1.53{\tiny\(\pm\textcolor{gray}{0.04}\)} & 3.33{\tiny\(\pm\textcolor{gray}{0.10}\)} & 0.19{\tiny\(\pm\textcolor{gray}{0.01}\)} & 0.52{\tiny\(\pm\textcolor{gray}{0.02}\)} \\
  & Whole-body RL            & 2.47{\tiny\(\pm\textcolor{gray}{0.05}\)} & 5.19{\tiny\(\pm\textcolor{gray}{0.19}\)} & \textcolor{myred}{\textbf{11.0}}{\tiny\(\pm\textcolor{gray}{0.3}\)} & 44.2{\tiny\(\pm\textcolor{gray}{1.0}\)} & 1.66{\tiny\(\pm\textcolor{gray}{0.03}\)} & 2.98{\tiny\(\pm\textcolor{gray}{0.09}\)} & \textcolor{myred}{\textbf{0.10}}{\tiny\(\pm\textcolor{gray}{0.01}\)} & \textcolor{myred}{\textbf{0.36}}{\tiny\(\pm\textcolor{gray}{0.01}\)} \\
  & \method                     & \textcolor{myred}{\textbf{1.48}}{\tiny\(\pm\textcolor{gray}{0.06}\)} & \textcolor{myred}{\textbf{4.78}}{\tiny\(\pm\textcolor{gray}{0.15}\)} & 11.4{\tiny\(\pm\textcolor{gray}{0.4}\)} & \textcolor{myred}{\textbf{42.3}}{\tiny\(\pm\textcolor{gray}{1.0}\)} & \textcolor{myred}{\textbf{0.33}}{\tiny\(\pm\textcolor{gray}{0.02}\)} & \textcolor{myred}{\textbf{1.97}}{\tiny\(\pm\textcolor{gray}{0.06}\)} & 0.14{\tiny\(\pm\textcolor{gray}{0.01}\)} & 0.39{\tiny\(\pm\textcolor{gray}{0.01}\)} \\

\midrule
\multirow{3}{*}{Push} 
  & Lower-body RL + IK       & 3.88{\tiny\(\pm\textcolor{gray}{0.16}\)} & 25.0{\tiny\(\pm\textcolor{gray}{1.2}\)} & 26.9{\tiny\(\pm\textcolor{gray}{1.4}\)} & \textcolor{myred}{\textbf{65.1}}{\tiny\(\pm\textcolor{gray}{4.2}\)} & 2.12{\tiny\(\pm\textcolor{gray}{0.10}\)} & 4.18{\tiny\(\pm\textcolor{gray}{0.16}\)} & 0.45{\tiny\(\pm\textcolor{gray}{0.06}\)} & 0.82{\tiny\(\pm\textcolor{gray}{0.07}\)} \\
  & Whole-body RL            & 4.62{\tiny\(\pm\textcolor{gray}{0.17}\)} & 29.6{\tiny\(\pm\textcolor{gray}{1.80}\)} & 20.1{\tiny\(\pm\textcolor{gray}{1.4}\)} & 70.3{\tiny\(\pm\textcolor{gray}{5.1}\)} & 2.38{\tiny\(\pm\textcolor{gray}{0.11}\)} & 6.20{\tiny\(\pm\textcolor{gray}{0.20}\)} & 0.91{\tiny\(\pm\textcolor{gray}{0.09}\)} & 1.83{\tiny\(\pm\textcolor{gray}{0.31}\)} \\
  & \method                     & \textcolor{myred}{\textbf{2.98}}{\tiny\(\pm\textcolor{gray}{0.10}\)} & \textcolor{myred}{\textbf{18.8}}{\tiny\(\pm\textcolor{gray}{0.65}\)} & \textcolor{myred}{\textbf{14.6}}{\tiny\(\pm\textcolor{gray}{0.9}\)} & 66.8{\tiny\(\pm\textcolor{gray}{3.6}\)} & \textcolor{myred}{\textbf{0.54}}{\tiny\(\pm\textcolor{gray}{0.05}\)} & \textcolor{myred}{\textbf{2.35}}{\tiny\(\pm\textcolor{gray}{0.12}\)} & \textcolor{myred}{\textbf{0.31}}{\tiny\(\pm\textcolor{gray}{0.05}\)} & \textcolor{myred}{\textbf{0.67}}{\tiny\(\pm\textcolor{gray}{0.06}\)} \\
\bottomrule
\end{tabular}
\endgroup}
\caption{Simulation Results: EE stability is evaluated in Isaac Gym across various tasks. \method consistently outperforms the baselines in most metrics, demonstrating superior EE stability.}
\label{sim_results}
\end{table*}
In this section, we evaluate the performance of \method in both simulation and real-world environments. Our experiments aim to answer the following key questions:
\begin{itemize}[leftmargin=*]
    \item \textbf{Q1} (Section~\ref{sec:sim}): Can the \emph{Two-Agent design} of \method perform better in \emph{simulation}? 
    \item \textbf{Q2}  (Section~\ref{sec:real}): What capabilities does \method{} enable in real world?
    \item \textbf{Q3} (Section~\ref{sec:freq}): How important is the \emph{Slow-Fast frequency design} for \method performance? 
\end{itemize}

\paragraph{Baselines.} We compare \method with the following baselines. 1) \textit{Robot Default Controller\footnote{This baseline is applicable only in the real world due to the accessibility of the built-in controller.}}~\cite{Unitree2024G1}: Utilizes the default Unitree locomotion, providing stable and low-impact locomotion. It serves as a naive baseline for EE stabilization. 2) \textit{Lower-body RL + IK}~\cite{ben2025homiehumanoidlocomanipulationisomorphic}: Employs a learned lower-body policy for locomotion followed by inverse kinematics to stabilize the EE. 3) \textit{Whole-body RL}: A single RL agent is trained to jointly control the whole body for both robust locomotion and stable EE control. 

\textbf{{Ablations of \method}.} We evaluate variants of \method{} using different upper-body and lower-body frequency pairings of 33.3 Hz, 50 Hz, and 100 Hz.

\vspace{-3pt}
\paragraph{Experiment Setup.}
We train our policy in Isaac Gym at 200 Hz simulation frequency. During training, reward functions, termination conditions, and curriculum design are consistent and frequency-agnostic across all comparisons. For real-world evaluation, we deploy \method on the Unitree G1 robot, following the sim-to-real pipeline of HumanoidVerse~\cite{humanoidverse}. o verify generalization, we transfer our framework to the Booster T1 robot~\cite{Booster2025T1} using the same frequency configuration(Visualization shown in Appendix~\ref{T1}). 

\vspace{-3pt}
\paragraph{Metrics.} We evaluated EE stability using the following metrics: linear acceleration norm (\emph{Acc}), angular acceleration norm (\emph{AngAcc}), and projected gravity in the XY plane of EE frame (\emph{Grav-XY}). Specifically, during locomotion, the z-direction may experience sudden velocity changes due to contact, so we additionally report the z-acceleration (\emph{Acc-Z}) for a more comprehensive evaluation. These metrics are reported as both the mean and maximum absolute values. Each metric is evaluated over 3 runs with mean and standard reported. For real-world acceleration, we collect pose data at 200 Hz using a mocap system for evaluation. The data is first interpolated, with abnormal points removed, and then double differentiation and filtering are applied to compute the acceleration. 

\subsection{Simulation Results}
\label{sec:sim}
To answer \textbf{Q1} (\textit{Can the \emph{Two-Agent design} of \method perform better in \emph{simulation}?}), we assess EE stability across three locomotion scenarios: (1) \emph{Tapping}: the robot steps in place to test stability under consistent, predictable contact events; (2) \emph{RandCommand}: where random commands are issued every 10 seconds to evaluate robustness across diverse motions; and (3) \emph{Push}: where the base is perturbed with a 0.5m/s velocity in a random direction every second to simulate unpredictable external disturbances. The results are summarized in Table~\ref{sim_results}.


We observe that \emph{Lower-body RL + IK} performs the worst due to lack of dynamics awareness, while \emph{Whole-body RL} improves but struggles to stabilize the EE in demanding scenarios like \emph{Push}, where external disturbances amplify the instability. In contrast, \method achieves the best overall performance, significantly reducing EE accelerations, especially in the vertical direction, highlighting the advantage of our decoupled design with frequency scheduling.

\begin{wrapfigure}{r}{0.4\linewidth}
    \centering
    \vspace{-15pt}
    \includegraphics[width=\linewidth]{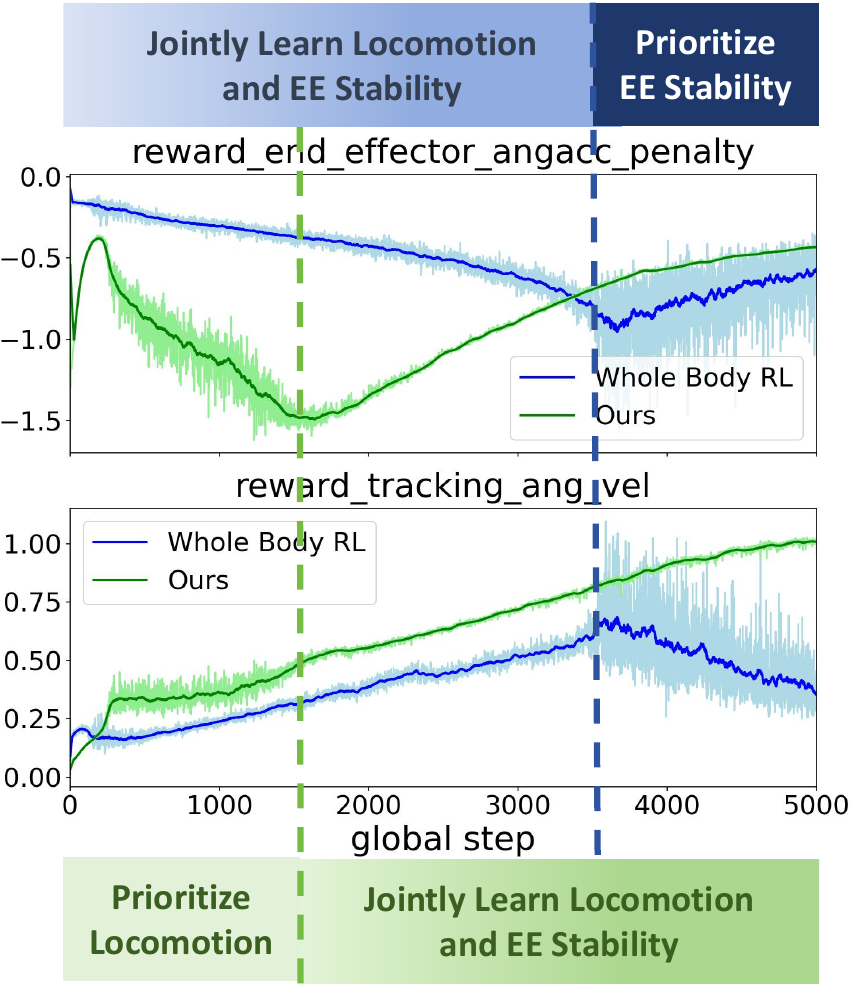}
    \caption{Reward Curves of EE-term and locomotion-term during Training.}
    \label{rw_conflict}
    \vspace{-25pt}
\end{wrapfigure}

\paragraph{Benefit from Two-Agent Reward Group Separation.} 

Figure~\ref{rw_conflict} shows the reward conflict between \texttt{EE-AngAcc-Penalty} for EE stabilization and \texttt{Angular-Vel-Tracking-Reward} for locomotion. For Whole-Body RL, optimizing both is difficult: prioritizing locomotion increases EE penalties, while a dominant EE penalty will make the RL to not keep standing all the time, sacrificing locomotion quality (see the last half of the blue line). In contrast, \method resolves this by decoupling the task objectives into two separate agents. Even with a significant EE penalty, the lower body keeps improving locomotion, then coordinates, enabling more stable learning and better performance.

\begin{figure}[h]
    \centering
    \includegraphics[width=0.85\linewidth]{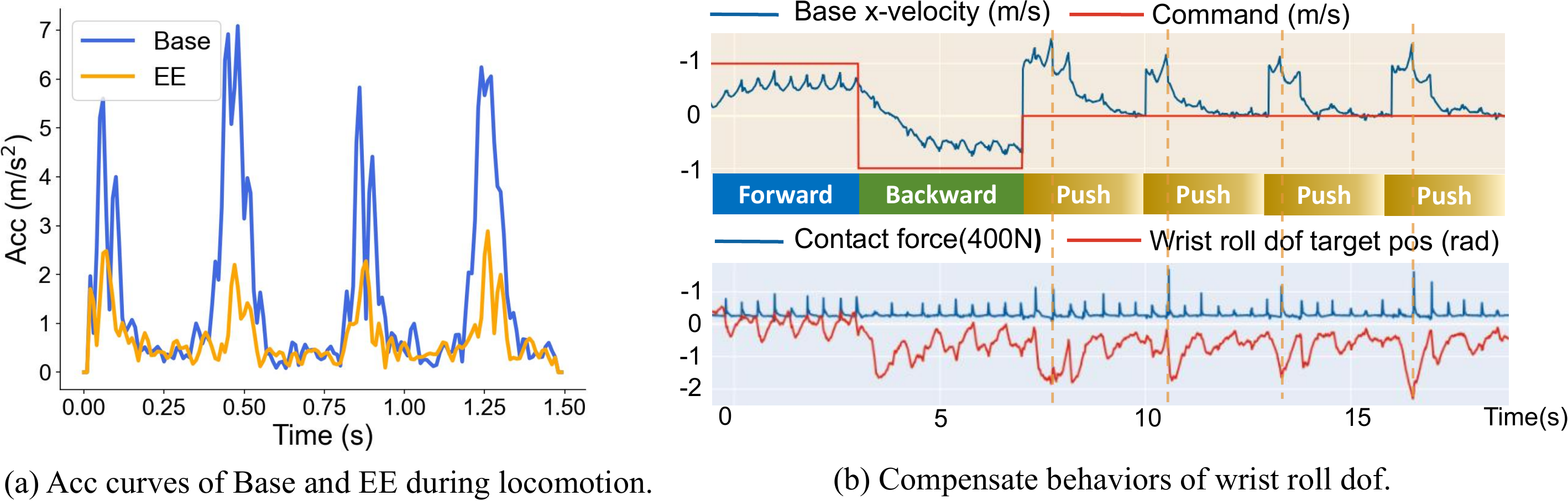}
    \caption{Emergent Compensation Behavior.}
    \label{Emergent_Com}
\end{figure}

\paragraph{Emergent Compensation Behavior.} Figure~\ref{Emergent_Com}(a) shows the acceleration curves of the base and EE. Our policy reduces sharp base accelerations caused by ground contacts, indicating stability through effective compensation, not just reduced base motion. To illustrate this, we visualize arm DoF target positions and contact force patterns. As seen in Figure~\ref{Emergent_Com}(b), DoF activations align with locomotion rhythm and contact events, with compensation peaking during external pushes and ground impacts, highlighting the upper body’s role in stabilizing the EE.

\vspace{-5mm}

\subsection{Real-World Results}
\label{sec:real}
To answer \textbf{Q2} (\textit{What capabilities does \method{} enable in real world?}), we assess EE stability in three real-world locomotion scenarios: (1) \emph{Tapping}; (2)\emph{TrajTrack} to move periodically along a straight line trajectory, and (3) \emph{Turning} to do in-place rotation. Note that the IK-based method relies heavily on motion capture system. Even with perfect state information in simulation, it fails to produce strong results, so we do not include it in the real-world experiments. 
 \begin{table*}[t]
\small
\vspace{-2mm}
\centering
\resizebox{\linewidth}{!}{%
\begingroup
\setlength{\tabcolsep}{2pt}
\renewcommand{\arraystretch}{0.8}
\begin{tabular}{llcccccccc}
\toprule
\multicolumn{2}{c}{Real-World Results} 
  & \multicolumn{2}{c}{Acc (m/s\textsuperscript{2}) \textcolor{myred}{\pmb{$\downarrow$}}}
  & \multicolumn{2}{c}{AngAcc (rad/s\textsuperscript{2}) \textcolor{myred}{\pmb{$\downarrow$}}} 
  & \multicolumn{2}{c}{Acc-Z (m/s\textsuperscript{2}) \textcolor{myred}{\pmb{$\downarrow$}}} 
  & \multicolumn{2}{c}{Grav-XY (m/s\textsuperscript{2}) \textcolor{myred}{\pmb{$\downarrow$}}} \\
\cmidrule(r){3-4} \cmidrule(r){5-6} \cmidrule(r){7-8} \cmidrule(r){9-10}
Task & Method 
  &  \phantom{0} mean  & \phantom{0} max \phantom{0} 
  & mean & max 
  &  \phantom{0}mean  &  \phantom{0} max  \phantom{0}
  & mean & max \\
\midrule
\multirow{3}{*}{Tapping} 
  & Robot Default Controller     
    & 4.67{\tiny\(\pm\textcolor{gray}{0.41}\)} & 9.71{\tiny\(\pm\textcolor{gray}{0.88}\)} 
    & 17.3{\tiny\(\pm\textcolor{gray}{2.0}\)} & 60.4{\tiny\(\pm\textcolor{gray}{6.8}\)} 
    & 1.25{\tiny\(\pm\textcolor{gray}{0.08}\)} & 4.01{\tiny\(\pm\textcolor{gray}{0.42}\)} 
    & 0.41{\tiny\(\pm\textcolor{gray}{0.05}\)} & 1.07{\tiny\(\pm\textcolor{gray}{0.10}\)} \\
  & Whole-body RL                
    & 1.86{\tiny\(\pm\textcolor{gray}{0.21}\)} & 6.11{\tiny\(\pm\textcolor{gray}{0.43}\)} 
    & 16.1{\tiny\(\pm\textcolor{gray}{1.5}\)} & 47.9{\tiny\(\pm\textcolor{gray}{5.3}\)} 
    & 1.34{\tiny\(\pm\textcolor{gray}{0.10}\)} & 5.10{\tiny\(\pm\textcolor{gray}{0.44}\)} 
    & \textcolor{myred}{\textbf{0.17}}{\tiny\(\pm\textcolor{gray}{0.02}\)} & 0.97{\tiny\(\pm\textcolor{gray}{0.11}\)} \\
  & \method                         
    & \textcolor{myred}{\textbf{1.35}}{\tiny\(\pm\textcolor{gray}{0.12}\)} & \textcolor{myred}{\textbf{4.96}}{\tiny\(\pm\textcolor{gray}{0.38}\)} 
    & \textcolor{myred}{\textbf{11.2}}{\tiny\(\pm\textcolor{gray}{1.1}\)} & \textcolor{myred}{\textbf{41.2}}{\tiny\(\pm\textcolor{gray}{5.5}\)} 
    & \textcolor{myred}{\textbf{0.52}}{\tiny\(\pm\textcolor{gray}{0.06}\)} & \textcolor{myred}{\textbf{2.33}}{\tiny\(\pm\textcolor{gray}{0.36}\)} 
    & 0.43{\tiny\(\pm\textcolor{gray}{0.02}\)} & \textcolor{myred}{\textbf{0.75}}{\tiny\(\pm\textcolor{gray}{0.08}\)} \\
\midrule
\multirow{3}{*}{TrajTrack} 
  & Robot Default Controller     
    & 4.88{\tiny\(\pm\textcolor{gray}{0.33}\)} & 11.6{\tiny\(\pm\textcolor{gray}{0.9}\)} 
    & 18.2{\tiny\(\pm\textcolor{gray}{4.1}\)} & 47.8{\tiny\(\pm\textcolor{gray}{5.0}\)} 
    & 1.41{\tiny\(\pm\textcolor{gray}{0.09}\)} & 5.53{\tiny\(\pm\textcolor{gray}{0.35}\)} 
    & 0.86{\tiny\(\pm\textcolor{gray}{0.06}\)} & 1.72{\tiny\(\pm\textcolor{gray}{0.15}\)} \\
  & Whole-body RL                
    & 2.95{\tiny\(\pm\textcolor{gray}{0.48}\)} & 12.2{\tiny\(\pm\textcolor{gray}{1.2}\)} 
    & 13.4{\tiny\(\pm\textcolor{gray}{1.4}\)} & 60.5{\tiny\(\pm\textcolor{gray}{8.7}\)} 
    & 2.02{\tiny\(\pm\textcolor{gray}{0.21}\)} & 9.51{\tiny\(\pm\textcolor{gray}{0.82}\)} 
    & 0.54{\tiny\(\pm\textcolor{gray}{0.04}\)} & 1.72{\tiny\(\pm\textcolor{gray}{0.11}\)} \\
  & \method                         
    & \textcolor{myred}{\textbf{1.51}}{\tiny\(\pm\textcolor{gray}{0.08}\)} & \textcolor{myred}{\textbf{6.25}}{\tiny\(\pm\textcolor{gray}{0.47}\)} 
    & \textcolor{myred}{\textbf{10.7}}{\tiny\(\pm\textcolor{gray}{0.7}\)} & \textcolor{myred}{\textbf{42.4}}{\tiny\(\pm\textcolor{gray}{3.3}\)} 
    & \textcolor{myred}{\textbf{0.62}}{\tiny\(\pm\textcolor{gray}{0.03}\)} & \textcolor{myred}{\textbf{3.17}}{\tiny\(\pm\textcolor{gray}{0.21}\)} 
    & \textcolor{myred}{\textbf{0.48}}{\tiny\(\pm\textcolor{gray}{0.03}\)} & \textcolor{myred}{\textbf{1.18}}{\tiny\(\pm\textcolor{gray}{0.09}\)} \\
\midrule
\multirow{3}{*}{Turning} 
  & Robot Default Controller    
    & 5.55{\tiny\(\pm\textcolor{gray}{0.28}\)} & 14.0{\tiny\(\pm\textcolor{gray}{0.4}\)} 
    & 23.2{\tiny\(\pm\textcolor{gray}{3.7}\)} & 62.1{\tiny\(\pm\textcolor{gray}{8.6}\)} 
    & 1.80{\tiny\(\pm\textcolor{gray}{0.09}\)} & 7.33{\tiny\(\pm\textcolor{gray}{0.37}\)} 
    & 0.90{\tiny\(\pm\textcolor{gray}{0.05}\)} & 1.83{\tiny\(\pm\textcolor{gray}{0.09}\)} \\
  & Whole-body RL                
    & 4.21{\tiny\(\pm\textcolor{gray}{0.21}\)} & 8.93{\tiny\(\pm\textcolor{gray}{0.45}\)} 
    & 16.2{\tiny\(\pm\textcolor{gray}{1.1}\)} & \textcolor{myred}{\textbf{57.9}}{\tiny\(\pm\textcolor{gray}{7.4}\)} 
    & 1.84{\tiny\(\pm\textcolor{gray}{0.11}\)} & 5.97{\tiny\(\pm\textcolor{gray}{0.30}\)} 
    & \textcolor{myred}{\textbf{0.31}}{\tiny\(\pm\textcolor{gray}{0.02}\)} & 0.87{\tiny\(\pm\textcolor{gray}{0.04}\)} \\
  & \method                         
    & \textcolor{myred}{\textbf{1.61}}{\tiny\(\pm\textcolor{gray}{0.08}\)} & \textcolor{myred}{\textbf{4.01}}{\tiny\(\pm\textcolor{gray}{0.20}\)} 
    & \textcolor{myred}{\textbf{9.41}}{\tiny\(\pm\textcolor{gray}{0.81}\)} & 62.8{\tiny\(\pm\textcolor{gray}{8.8}\)} 
    & \textcolor{myred}{\textbf{0.72}}{\tiny\(\pm\textcolor{gray}{0.04}\)} & \textcolor{myred}{\textbf{3.94}}{\tiny\(\pm\textcolor{gray}{0.20}\)} 
    & 0.36{\tiny\(\pm\textcolor{gray}{0.02}\)} & \textcolor{myred}{\textbf{0.71}}{\tiny\(\pm\textcolor{gray}{0.04}\)} \\

\bottomrule
\end{tabular}
\endgroup}
\caption{Real-World Results: EE stability evaluated in Real World across diverse task settings. \method{} consistently outperforms baselines, especially in Acc-Z metric.}
\label{real_results}
\end{table*}

The results in Table~\ref{real_results} show that the \emph{Robot Default Controller} exhibits the highest acceleration across nearly all metrics, highlighting that even carefully designed locomotion controllers with gentle stepping are insufficient for tasks requiring precise EE stability. While \emph{Whole-body RL} offers moderate improvements, it struggles under motions with large movement like \emph{TrajTrack}. In contrast, \method{} maintains consistent and robust performance even during diverse locomotion. Compared to simulation results, real-world tests reveal that despite using the same domain randomization, observation noise, and reward functions, \method demonstrates stronger sim-to-real transferability. Whole-body RL, by comparison, shows noticeably sluggish and hesitant steps, with shifts during foot tapping, likely due to excessive upper-body influence.

With the EE stability plus robust locmotion, \method{} enables the robot to perform the following precise and stable upper-body tasks during locomotion. 

\begin{figure}
    \centering
    \includegraphics[width=0.87\linewidth]{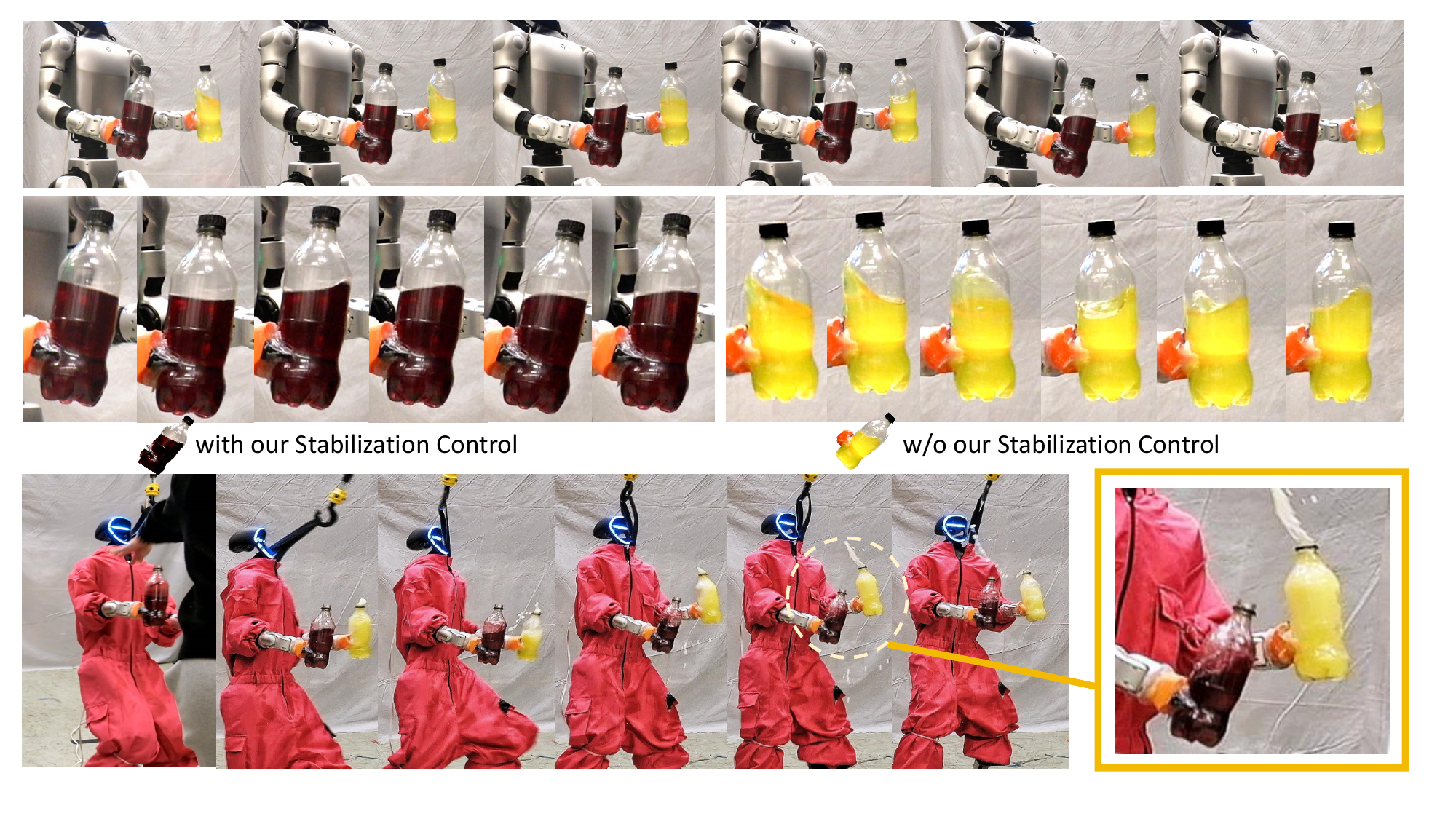}
    \caption{\emph{Top:} Humanoid carring bottle of water without spillage during tepping. \emph{Bottom:} Humanoid disturbance rejection with EE stability.}
    \label{fig:liquid_comparison}
      \vspace{-5pt}
\end{figure}

\textbf{Case 1: Humanoid Carrying Bottle without Spillage.} Figure~\ref{fig:liquid_comparison} shows the humanoid carrying a water bottle during locomotion. Even in tapping, without stabilization (\textcolor{baselinecolor}{\textbf{YELLOW}}, no compensation behavior encouraged), contact impacts cause the liquid to slosh noticeably. In contrast, \method{} (\textcolor{ourcolor}{\textbf{RED}}) greatly suppresses liquid motion, allowing the robot to carry an almost full cup of water smoothly while walking. Beyond periodic locomotion, our policy also demonstrates strong disturbance rejection capabilities. As shown in the Figure~\ref{fig:liquid_comparison}, when subjected to sudden and forceful pushes, the robust locomotion of the robot quickly adapts to avoid falling, while the upper body actively compensates to keep the end effector as steady as possible, effectively preventing the liquid from spilling.

\begin{figure}[ht]
    \centering
    \vspace{-5pt} 
    \includegraphics[width=0.8\linewidth]{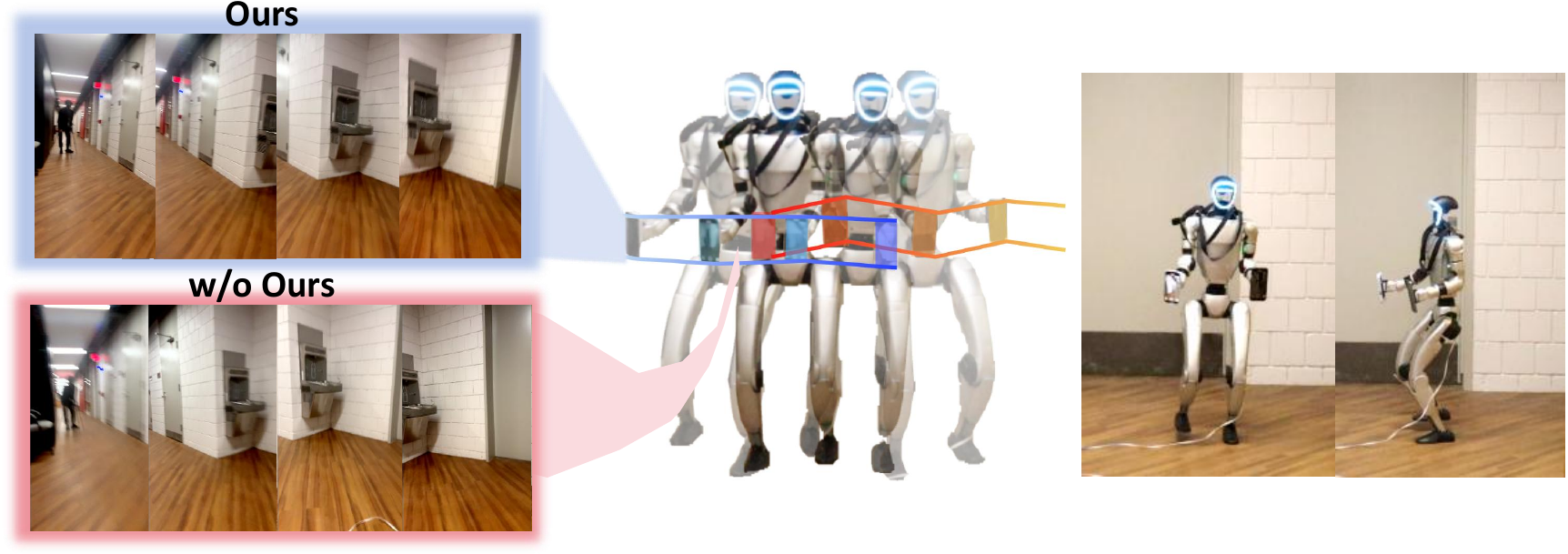}
    \caption{Humanoid as Camera Stabilizer to record videos.}
    \label{fig:camera_comparison}
    \vspace{-5pt} 
\end{figure}

\textbf{Case 2: Humanoid as Camera Stabilizer.} Figure~\ref{fig:camera_comparison} shows video footage recorded by the robot during continuous turning, comparing with and without stabilization. \method{} ensures smooth and consistent camera motion, avoiding visible jitter with off-gantry-level robust locomotion. This allows the robot to record long, uninterrupted videos.

\begin{wraptable}{r}{0.5\linewidth}
\vspace{-22pt}
\raggedleft
\small
\resizebox{\linewidth}{!}{%
\begingroup
\setlength{\tabcolsep}{4pt}
\renewcommand{\arraystretch}{1.0}
\begin{tabular}{llcccccc}
\toprule
\multicolumn{2}{c}{} 
  & \multicolumn{2}{c}{IsaacGym} 
  & \multicolumn{2}{c}{Sim2Sim} 
  & \multicolumn{2}{c}{Sim2Real} \\
\cmidrule(r){3-4} \cmidrule(r){5-6} \cmidrule(r){7-8}
\multicolumn{2}{c}{Acc (m/s\textsuperscript{2}) \textcolor{myred}{\pmb{$\downarrow$}}} 
  & mean & max 
  & mean & max 
  & mean & max \\
\midrule
\multirow{1}{*}{w/o Tracking}    
  & & \textcolor{myred}{\textbf{0.82}} & \textcolor{myred}{\textbf{2.06}} & 2.29 & 7.83 & 2.91 & 8.15 \\
\multirow{1}{*}{with Tracking} 
  & & 1.08 & 3.45 & \textcolor{myred}{\textbf{1.14}} & \textcolor{myred}{\textbf{4.05}} & \textcolor{myred}{\textbf{1.35}} & \textcolor{myred}{\textbf{4.96}} \\
\bottomrule
\end{tabular}
\endgroup}
\vspace{-2pt}
\caption{EE Tracking Improve Transferability}
\vspace{-15pt}
\label{tab:tracking_vs_fixed}
\end{wraptable}

\paragraph{Benefits of Random EE Position Command for Sim2Real.} While both the whole-body RL and \method{} incorporate EE tracking, we found that tracking, specifically by forcing the EE to remain stable at a given position, plays a crucial role in generalization. Fixing the EE position during training often led to overfitted behaviors tied to specific poses and simulator dynamics, resulting in poor transferability. In contrast, training with random EE position commands promotes more reactive and adaptable motions, fostering compensation patterns that transfer more effectively (Table~\ref{tab:tracking_vs_fixed}).

\subsection{In-Depth Analysis on Frequency Design}
\label{sec:freq}
\begin{figure}[ht]
    \centering
    \includegraphics[width=0.9\linewidth]{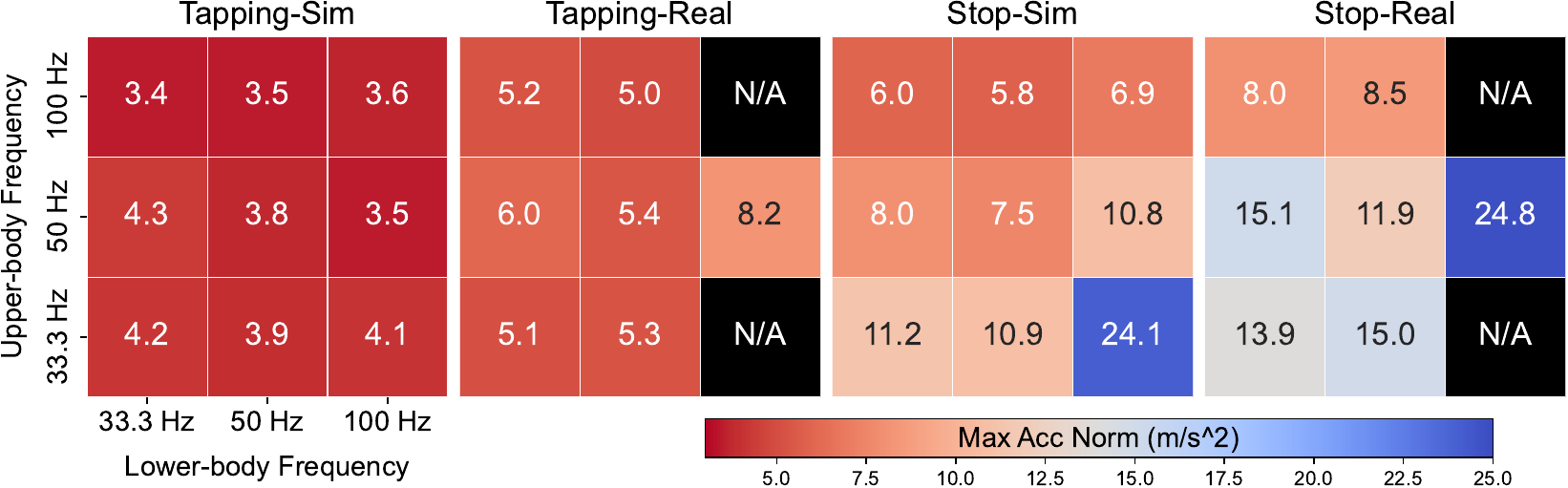}
\caption{Max Acc under Different Control Frequencies in Simulation and Real World: \textcolor{myblue}{Higher values} reflect reduced stability. N/A indicates unstable or failed trials in the real-world testing.}
\vspace{-5pt}
    \label{fig:freq_abl}
\end{figure}

To answer \textbf{Q3} (\textit{How important is the \emph{Slow-Fast frequency design} for \method performance?}), we compare peak EE acceleration under various frequency settings in both simulation and real-world environments, using two scenarios: \emph{Tapping} (predictable contacts) and \emph{Stop} (sudden switch from walking to standing). As shown in Figure~\ref{fig:freq_abl}, our slow-fast design, 50 Hz for the lower body and 100 Hz for the upper, consistently achieves lower acceleration across tasks and domains.

We observe that in simulation, a 50 Hz lower-body policy is sufficient for maintaining stable locomotion, even under unpredictable conditions. In contrast, a higher-frequency upper-body policy proves beneficial for rapid recovery during sudden stops. From a sim-to-real perspective, the results indicate that deploying a high-frequency lower-body policy may introduce stricter deployment constraints. In our real-\emph{Stop} trials, a 100 Hz lower-body policy caused more oscillations and degraded EE performance, with some instances resulting in failure. This degradation may be due to increased sensitivity to observation noise and control delays. On the other hand, running the upper-body agent at 100 Hz did not exhibit such issues and consistently enhanced overall performance. Considering that 50 Hz locomotion is a widely adopted standard and that inference-time constraints (0.01 s) are in place, our Slow-Fast Frequency configuration appears to be near-optimal. The further analysis of high-frequency upper body behaviors can be shown in Appendix~\ref{freq_analysis}.

\subsection{Cross Embodiment Validation}
\vspace{-5pt}
\label{T1}

To assess the generalizability of our policy training methodology and control design, we perform cross-embodiment validation by applying the same training procedure to a distinct robot embodiment, Booster T1~\cite{Booster2025T1}, which differs from the base humanoid in joint configuration and body proportions. This evaluation focuses on whether the design principles—such as the Slow-Fast Frequency strategy—lead to consistently effective behavior when applied to a new morphology.

\begin{figure}
    \centering
    \includegraphics[width=0.85\linewidth]{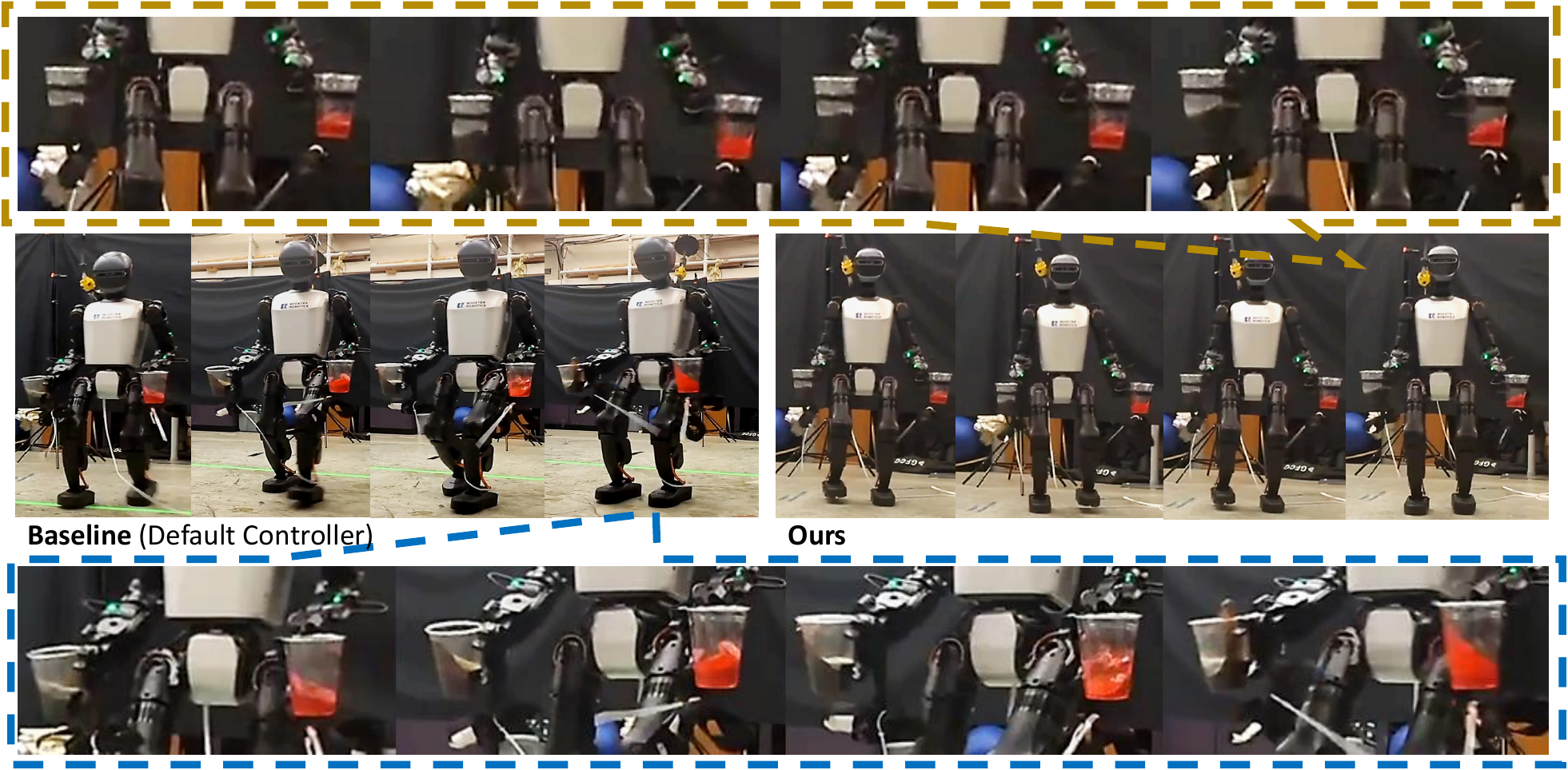}
    \caption{Real-world Results on Booster T1. The right hand holding the cola is controlled by our stabilization controller.}
    \label{rw_conflict}
\end{figure}

Despite differences in embodiments, the T1 policy trained with the same design framework exhibits better stability and coordination than its Default Controller, particularly in tasks that involve sudden stops and precise end-effector control. These results suggest that our training approach captures transferable structural priors that support robust behaviors across diverse humanoid platforms, without requiring embodiment-specific tuning.

\section{Conclusion}
In this paper, we present \method, a Slow-False Two-Agent reinforcement learning framework that enables robust locomotion and precise, stable EE control through frequency separation and task-specific reward design. Extensive experiments show up \method can have a 50-80\% reduction in EE acceleration, achieving much more closer to human-level stability. This allows successful deployment of tasks like walking while carrying liquids or recording stable video on the Unitree G1 humanoid and enabling humanoid robots to perform complex tasks with precision and reliability.

\section{Limitation}

Despite its strong performance, \method{} still faces several limitations. First, while it significantly reduces EE acceleration, the achieved stability still falls short of human-level performance. Carrying a cup of water while walking is a task that humans can perform effortlessly with minimal spill. \method{} yet match the subtlety and adaptability of human control. Second, the decoupling of locomotion and end-effector control creates a fixed task boundary. While this separation is effective for many loco-manipulation tasks, it becomes suboptimal when the two modules must closely coordinate, such as during dynamic reaching or complex interactions. Third, while \method{} offers a flexible framework for many scenarios and introduces valuable insights on frequency assignment, its performance may vary depending on the specific task or robot configuration. Aspects like task complexity, robot morphology, or the need for more nuanced coordination may require further adjustments to the design.

Future work could focus on improving the adaptability of \method{} to more diverse tasks and robot configurations, with particular attention to dynamic coordination and complex interactions. Additionally, addressing the human-level stability gap will be crucial, particularly in tasks requiring high precision and fine motor control. Exploring more advanced learning strategies and architecture, such as attention mechanism, could help achieve better generalization across various platforms and tasks.


\clearpage
\acknowledgments{We sincerely thank Zhengyi Luo and Zenghao Tang for their insightful discussions, which significantly contributed to shaping the direction of this work. We also thank Haotian Lin and Nikhil Sobanbabu for their help with real-world experiments. We also gratefully acknowledge the hardware support provided by Unitree Robotics and Booster Robotics.}


\bibliography{main}  

\clearpage
\appendix
\section{Appendix}

\subsection{Training Details}
\label{appendix:training}

\paragraph{Observation}
We adopt an asymmetric observation structure to enable efficient policy learning in simulation while ensuring robust real-world deployment under partial observability. The actor relies solely on onboard-accessible inputs—proprioception, command signals, and recent actions—excluding global position data, thus removing dependence on odometry or external tracking. Observations are stacked over five timesteps to provide short-term temporal context.

\begin{table*}[ht]

\centering
\begin{tabular}{llcccc}
\toprule
\textbf{Type} & \textbf{Observation} & \textbf{Actor} & \textbf{Critic} & \textbf{Scale}  & \textbf{Noise Scale} \\
\midrule
\multirow{3}{*}{\textbf{Privileged}} 
  & base\_lin\_vel               & {\color{red}\ding{55}} & \ding{51} & 2.0 & 0.0 \\
  & end\_effector\_relative\_pos & {\color{red}\ding{55}} & \ding{51} & 1.0 & 0.0 \\
  & end\_effector\_gravity       & {\color{red}\ding{55}} & \ding{51} & 1.0 & 0.0 \\
    \arrayrulecolor{gray!50}
  \midrule
\multirow{6}{*}{\textbf{Proprioception}} 
  & base\_ang\_vel              & \ding{51} & \ding{51} & 0.25 & 0.1 \\
  & projected\_gravity         & \ding{51} & \ding{51} & 1.0  & 0.0 \\
  & dof\_pos                   & \ding{51} & \ding{51} & 1.0  & 0.01 \\
  & dof\_vel                   & \ding{51} & \ding{51} & 0.05 & 0.1 \\
  & actions                    & \ding{51} & \ding{51} & 1.0  & 0.0 \\
  & sin\_phase / cos\_phase    & \ding{51} & \ding{51} & 1.0  & 0.0 \\
  \arrayrulecolor{gray!50}
  \midrule
  \multirow{5}{*}
{\textbf{Command}} 
  & command\_lin\_vel          & \ding{51} & \ding{51} & 1.0 & 0.0 \\
  & command\_ang\_vel          & \ding{51} & \ding{51} & 1.0 & 0.0 \\
  & command\_EE                & \ding{51} & \ding{51} & 1.0 & 0.0 \\
  & command\_gait              & \ding{51} & \ding{51} & 1.0 & 0.0 \\
  \arrayrulecolor{black}
  \bottomrule
\end{tabular}
\caption{Comparison of actor and critic observations with scaling factors. Privileged observations used only by the critic are shaded and marked in red.}
\label{tab:observations}
\end{table*}

During training, the critic is granted privileged access to additional information, including \texttt{base\_lin\_vel}, \texttt{end\_effector\_relative\_pos}, and \texttt{end\_effector\_gravity}, which help robot to understand its current state and task success more accurately. To improve robustness, noise is injected into selected observations. Observation scales and noise scales are summarized in Table~\ref{tab:observations}.

This setup improves value estimation and training stability while ensuring deployable policies grounded in realistic sensor inputs, supporting robust sim-to-real transfer for locomotion and end-effector tasks.

\paragraph{Task Definition}

We define our task as a combination of robust locomotion and end-effector (EE) stabilization under general body configurations. The EE stabilization command, denoted as \(c^{\text{EE}} \in \mathbb{R}^5\), encodes task-specific requirements. The first dimension is a binary flag indicating whether EE stabilization is enabled. If this value is zero, all stabilization-related rewards are disabled for that EE. The next two values specify the desired EE position \((x, y)\) in the local frame of the body. The fourth value defines the target EE height along the global \(z\)-axis, given as an offset relative to the desired base \(z\)-position. The final element of \(\text{c}^{\text{EE}}\) is a tolerance parameter $\sigma_{EE}$ that controls the precision of EE tracking. A higher tolerance leads to smoother motion with lower accelerations, which is beneficial for tasks such as bottle carrying where precise EE positioning is not critical. Conversely, a lower tolerance prioritizes accurate tracking, which is essential for tasks like camera stabilization where EE pose must be tightly maintained.

For locomotion, the control command includes both target base velocity and gait information. The velocity command comprises desired linear velocities \((v_x, v_y)\) and angular velocity \(\omega\), all defined in the base frame. The system is expected to track these velocities within specified tolerances \(\sigma_x, \sigma_y, \sigma_\omega\). Gait control is represented by a two-dimensional vector. The first value is a binary indicator of whether the desired gait is a double-stance (both feet in contact). If not (i.e., in dynamic gait mode), the second value specifies the desired gait period. From this gait period, we compute the gait phase using sinusoidal signals \((\sin(\phi), \cos(\phi))\), where \(\phi\) denotes the phase. This allows the derivation of target contact timings for each foot $\hat{C}$. A phase-based reward is then introduced to guide the agent to follow the desired contact sequence $\hat{C}$.

We list all command ranges in Table~\ref{tab:command_ranges}, with $\sigma_x=0.5 \text{m/s}$, $\sigma_y=0.5 \text{m/s}$, $\sigma_{\omega}=0.5 \text{rad/s}$ respectively.

\begin{table*}[ht]
\centering
\begin{tabular}{lc}
\toprule
\textbf{Component} & \textbf{Range / Value} \\
\midrule
\multirow{2}{*}{\textbf{\texttt{command\_lin\_vel}}} & x: $\mathcal{U}$ (-1, 1) m/s  \\
  & y: $\mathcal{U}$ (-1, 1) m/s \\
  \arrayrulecolor{gray!50}
  \midrule
  \textbf{\texttt{command\_ang\_vel}}
  & $\mathcal{U}$ (-1, 1) rad/s  \\
  \midrule
\multirow{2}{*}{\textbf{\texttt{command\_gait}}} & mode: 0/1\\
  & period: $\mathcal{U}$ (0.5, 1.3) m/s \\
\midrule
\multirow{5}{*}{\textbf{\texttt{command\_EE}}} 
&activation: 0/1 \\
& x: $\mathcal{U}$ (-0.15, 0.15) m  \\
  & y: $\mathcal{U}$ (-0.15, 0.15) m \\
    & z: $\mathcal{U}$ (-0.2, 0.2) m \\
    & tolerance: $\mathcal{U}$ (0.1, 0.2) m\\
    \arrayrulecolor{black}
\bottomrule
\end{tabular}
\caption{Command ranges used during training.}
\label{tab:command_ranges}
\end{table*}

\paragraph{Domain Randomization}
To enhance the robustness and generalization of \method{}, we apply domain randomization techniques, as detailed in Table~\ref{tab:domain_randomization}. We first train \method{} with all domain randomization strategies listed, excluding push perturbations. After obtaining a stable policy, we introduce push disturbances to further improve robustness under external disturbance.
\begin{table*}[ht]
\centering
\begin{tabular}{lc}
\toprule
\textbf{Component} & \textbf{Range / Value} \\
\midrule
\textbf{P Gain}
  & $\mathcal{U}$ (0.95, 1.05) $\times$ \texttt{default}  \\
  \arrayrulecolor{gray!50}
  \midrule
  \textbf{D Gain}
  & $\mathcal{U}$ (0.95, 1.05) $\times$ \texttt{default}  \\
  \midrule
\textbf{Friction Coefficient} & $\mathcal{U}$ (-0.5, 1.25) \\
\midrule
\textbf{Base Mass} & $\mathcal{U}$ (-1.0, 3.0)  \texttt{kg}\\
\midrule
\textbf{Control Delay} & $\mathcal{U}$ (20, 40)  \texttt{ms} \\
\midrule
\multirow{2}{*}{\textbf{Push Perturbations}} & \texttt{Interval:} $\mathcal{U}$ (5, 16) \texttt{s} \\
                                                                & \texttt{Max velocity:} 0.5 \texttt{m/s} \\
  \midrule
\multirow{4}{*}{\textbf{External Force on EE}} 
  & \texttt{Position std:} 0.03  \texttt{m} \\
  & \texttt{X force:}  $\mathcal{U}$ (-0.5, 0.5)  \texttt{N}\\
  & \texttt{Y force:}  $\mathcal{U}$ (-0.5, 0.5) \texttt{N}\\
  & \texttt{Z force:}  $\mathcal{U}$ (-7, 2) \texttt{N}\\
    \arrayrulecolor{black}
\bottomrule
\end{tabular}
\caption{Domain randomization parameters used during training.}
\label{tab:domain_randomization}
\end{table*}

\paragraph{Rewards Design}
We show the grouped \method{} task reward components in Table~\ref{rw_conflict}. Notice that the termination is a shared reward component Also, we introduce several penalties and energy regularization in order to achieve robust sim-to-real performance like \emph{dof limit, stand symmetry, contact force, feet height on the air, action rate} and so on. Follow ~\cite{asap}, We adjust the scaling factor \( s_{t,i} \) in the cumulative discounted reward formula to handle small rewards differently based on their sign: $\mathbb{E}\left[\sum_{t=1}^T \gamma^{t-1} \sum_{i} s_{t,i} r_{t,i}\right],$
where \( s_{t,i} = s_{\text{current}} \) if \( r_{t,i} < 0 \), and 1 if \( r_{t,i} \geq 0 \). The factor \( s_{\text{current}} \) starts at 0.5 and is adjusted dynamically—multiplied by \( 0.9999 \) when episode length is under 0.4s, and by \( 1.0001 \) when it exceeds 2.1s, with an upper bound of 1. This allows our policy to first focus on task terms and them regular the behavior to be smooth and reasonable for sim-to-real.

\begin{table*}[ht]
\label{tab:reward_groups}
\small
\centering
\begin{tabular}{ l  l  c  l }
\hline
Group & Term & Weight & Expression \\
\hline
\multirow{6}{*}{\textbf{Lower Body}} 
& \texttt{tracking\_lin\_vel\_x} & 1.5 & $\exp(-\frac{1}{\sigma_x^2}\lVert v_x - \hat{v}_x \rVert^2)$ \\
& \texttt{tracking\_lin\_vel\_y} & 1.0 & $\exp(-\frac{1}{\sigma_y^2}\lVert v_y - \hat{v}_y \rVert^2)$ \\
& \texttt{tracking\_ang\_vel} & 2.0 & $\exp(-\frac{1}{\sigma_{\omega}^2}\lVert \omega_z - \hat{\omega}_z \rVert^2)$ \\
& \texttt{tracking\_base\_height} & 0.5 & $\exp(-\frac{1}{\sigma_h}\lVert h - \hat{h} \rVert)$ \\
& \texttt{tracking\_gait\_contact} & 0.5 & $\sum (\mathds{1}(C=\hat{C})- \mathds{1}(C\neq \hat{C}))$ \\
& \texttt{termination} & -500.0 & $\mathds{1}_{\text{terminate}}$ \\
\midrule
\multirow{7}{*}{\textbf{Upper Body}} 
& \texttt{tracking\_end\_effector\_pos} & 1.0  
& $\exp\left(-\frac{1}{\sigma_{EE}^2} \| p_{\text{EE}} - \hat{p}_{\text{EE}} \|^2 \right)$ \\

& \texttt{tracking\_zero\_end\_effector\_acc} & 10 
& $\exp\left(-\lambda_{\text{acc}} \| \ddot{p}_{\text{EE}} \|^2 \right)$ \\

& \texttt{tracking\_zero\_end\_effector\_ang\_acc} & 1.5 
& $\exp\left(-\lambda_{\text{ang-acc}} \| \dot{\omega}_{\text{EE}} \|^2 \right)$ \\

& \texttt{penalty\_end\_effector\_acc} & -0.1 
& $- \| \ddot{p}_{\text{EE}} \|^2$ \\

& \texttt{penalty\_end\_effector\_ang\_acc} & -0.01 
& $- \| \dot{\omega}_{\text{EE}} \|^2$ \\

& \texttt{penalty\_end\_effector\_tilt} & -5.0 
& $- \left\| \mathbf{P}_{xy}(R_{\text{EE}}^T \mathbf{g}) \right\|^2$ \\
& \texttt{termination} & -100.0 & $\mathds{1}_{\text{terminate}}$ \\

\hline
\end{tabular}
\caption{Reward terms categorized by body group, including task rewards and penalties with corresponding expressions and weights. $C$ means the contact sequence. Hat over variables represents the desired value. In implementation, we set $\lambda_{acc}=0.25$, $\lambda_{acc}=0.0044$.}
\label{tab:rewards}
\end{table*}

\paragraph{Training Hyperparameter}
We summarize the main hyperparameters used in our PPO multi-actor-critic training setup in Table~\ref{tab:ppo_config}. These include general PPO settings, action std for different body modules, and the network architecture shared across policy and value networks.
\begin{table*}[ht]
\centering

\begin{tabular}{ l l }
\hline
\textbf{Parameter} & \textbf{Value} \\
\hline
\rowcolor{gray!30}
\multicolumn{2}{c}{\textit{General PPO Settings}} \\

Gamma (\( \gamma \)) & 0.99 \\
GAE Lambda (\( \lambda \)) & 0.95 \\
Value Loss Coef & 1.0 \\
Entropy Coef & 0.01 \\
Actor Learning Rate & \(1 \times 10^{-3}\) \\
Critic Learning Rate & \(1 \times 10^{-3}\) \\
Max Grad Norm & 1.0 \\
Desired KL & 0.01 \\
Num Steps per Env & 48 \\

\rowcolor{gray!30}
\multicolumn{2}{c}{\textit{Noise Settings}} \\
Init Noise Std & lower\_body: 0.8, upper\_body: 0.6 \\
Std Threshold & lower\_body: 0.15, upper\_body: 0.10 \\
\rowcolor{gray!30}
\multicolumn{2}{c}{\textit{Network Architecture}} \\
Hidden Layers & [512, 256, 128] \\
Activation Function & ELU \\
\hline
\end{tabular}
\caption{PPO Multi-Actor-Critic Training Configuration}
\label{tab:ppo_config}
\end{table*}

\subsection{More Analysis on Frequency Ablation}
\label{freq_analysis}
\begin{figure}[h]
    \centering
        \begin{subfigure}[t]{0.7\textwidth}
        \centering
        \includegraphics[width=\linewidth]{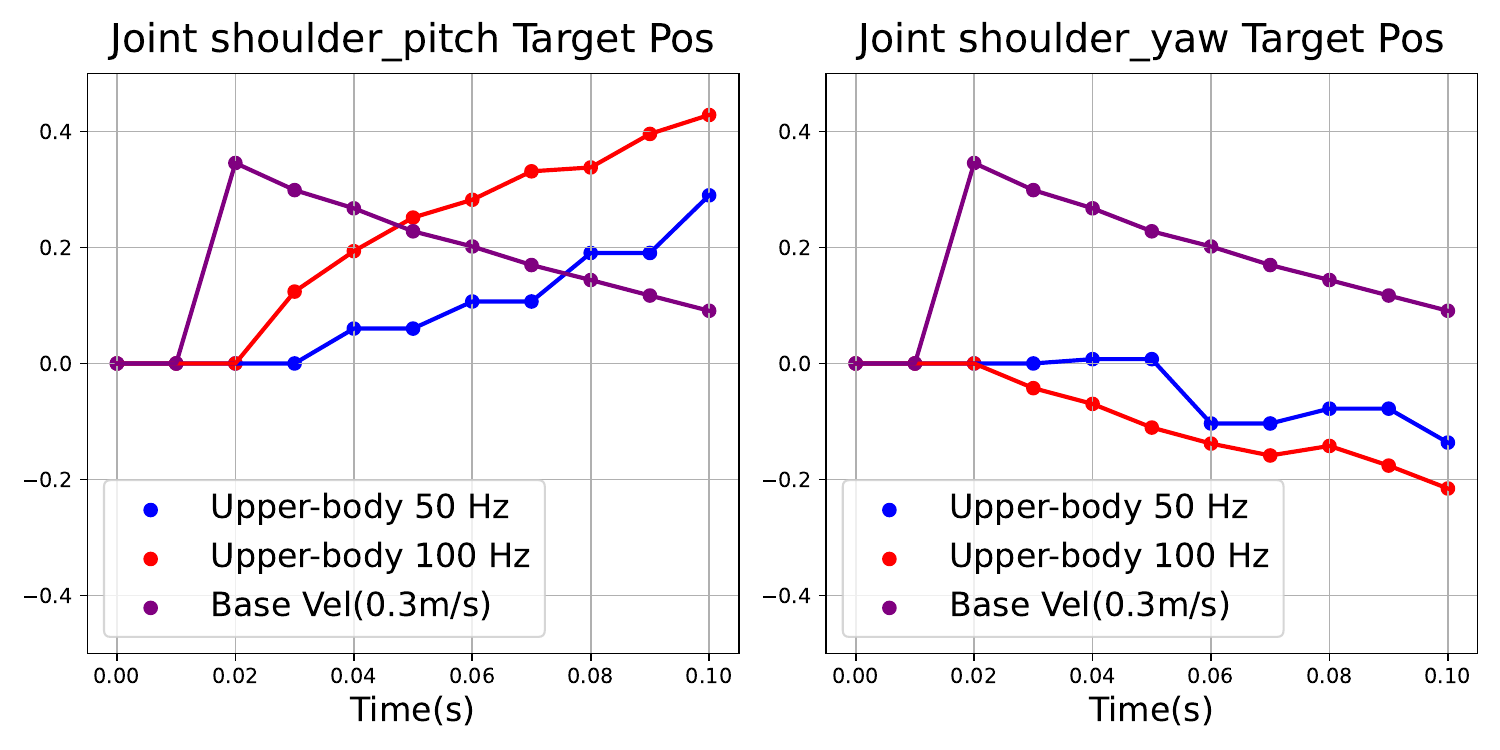}
    \end{subfigure}
    \begin{subfigure}[t]{0.6\textwidth}
        \centering
        \includegraphics[width=\linewidth]{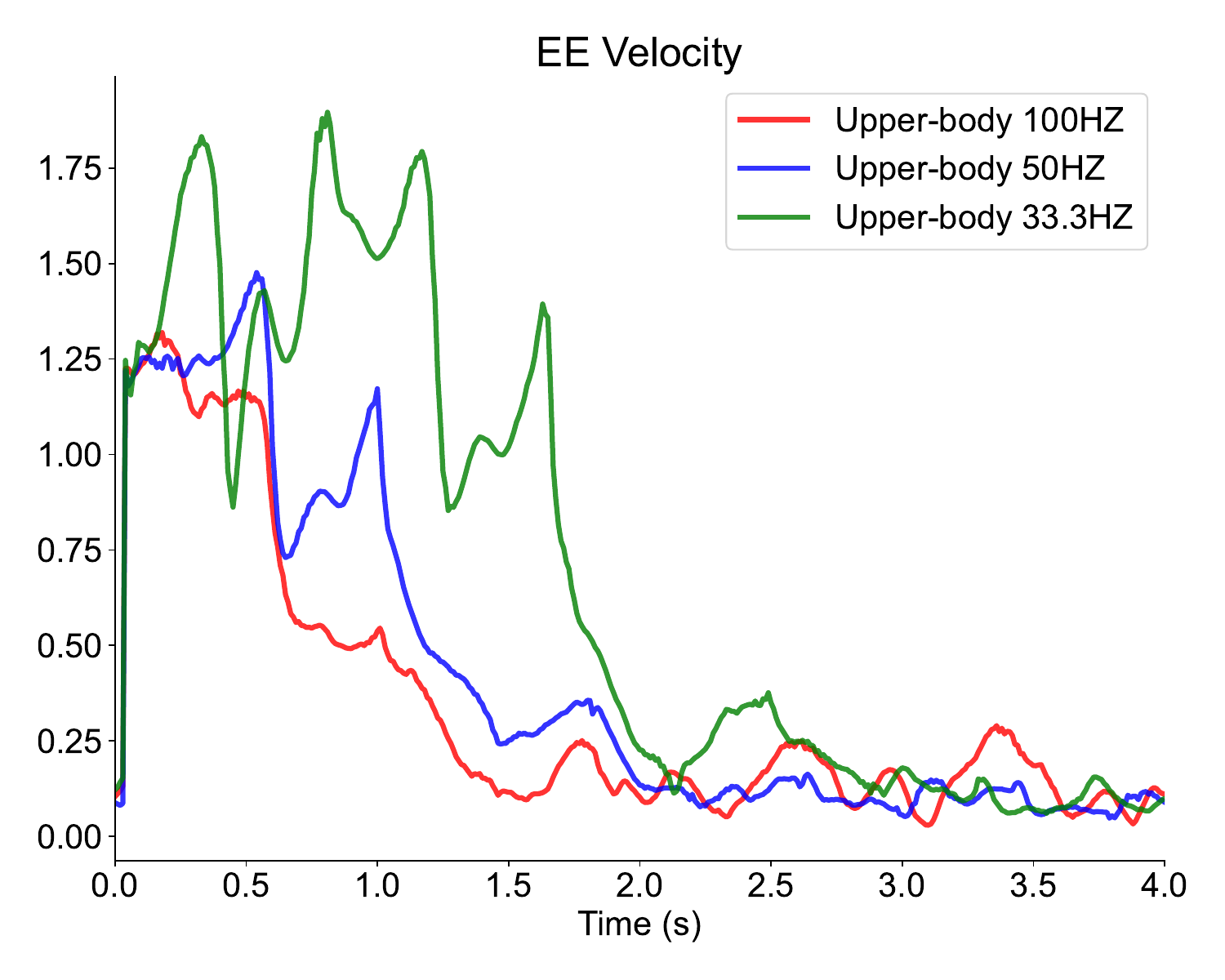}
    \end{subfigure}

    \caption{Effect of upper-body control frequency on EE stabilization. top: EE velocity (m/s) recovery with different upper-body frequencies. bottom: Response comparison at 100 Hz vs. 50 Hz.}
    \label{fig:freq_ablation}
\end{figure}

\begin{table*}[h]
\centering
\small
\resizebox{0.8\linewidth}{!}{%
\begingroup
\setlength{\tabcolsep}{4pt}
\renewcommand{\arraystretch}{1.0}
\begin{tabular}{lccc}
\toprule
Methods & Response Time (s) \textcolor{myred}{\pmb{$\downarrow$}} & Max Acc (m/s\textsuperscript{2})  \textcolor{myred}{\pmb{$\downarrow$}} & Max Vel (m/s) \textcolor{myred}{\pmb{$\downarrow$}}\\
\midrule
Ours (L50-U33)  & 0.598 & 43.5 & 1.90 \\
Ours (L50-U50)  & 0.338 & 40.5 & 1.48 \\
Ours (L50-U100) & \textcolor{myred}{\textbf{0.167}} & \textcolor{myred}{\textbf{37.8}} & \textcolor{myred}{\textbf{1.17}} \\
\bottomrule
\end{tabular}
\endgroup}
\caption{Response time and maximum error magnitudes under different upper-body frequencies.}
\label{tab:freq_ablation}
\end{table*}
Across both simulation and real-world environments, our experiments show that a 50 Hz lower-body control frequency consistently achieves stable locomotion, regardless of the upper-body control frequency, whereas other lower-body frequencies may lead to degraded performance under certain upper-body control frequencies. We further investigated how higher upper-body frequencies enhance EE stability in challenging scenarios, such as sudden external pushes. As shown in Figure~\ref{fig:freq_ablation} (top), higher-frequency policies (100 Hz) react faster to base motion changes and recover balance quicker. In Figure~\ref{fig:freq_ablation} (bottom), we observe that higher frequencies lead to faster EE velocity recovery. Table~\ref{tab:freq_ablation} presents the quantitative results. We observe that increasing the upper-body control frequency reduces recovery time (defined as the time when the error first falls below $\frac{1}{e}$ of its maximum), as well as peak acceleration and velocity errors. This indicates enhanced disturbance compensation and faster recovery dynamics.

\end{document}